\pdfoutput=1

\documentclass[11pt]{article}
\usepackage[subtle]{savetrees}

\usepackage{emnlp2021}

\usepackage{times}
\usepackage{latexsym}

\usepackage[T1]{fontenc}

\usepackage[utf8]{inputenc}

\usepackage{microtype}

\usepackage{booktabs}
\usepackage{multirow}

\usepackage{graphicx}
\usepackage{amssymb}
\usepackage{titlesec}
\titlespacing*{\section}{0pt}{4pt}{4pt}
\titlespacing*{\subsection}{0pt}{2pt}{4pt}

\title{ReInform: Selecting paths with reinforcement learning \\for contextualized link prediction}

\author{Marina Speranskaya \\
  Center for Information and \\ 
  Language Processing, LMU \\
  Munich, Germany \\
  \texttt{speranskaya@cis.lmu.de} \\\And
  Sameh Methias \\
  Technical University of Munich \\
  Munich, Germany \\
  \texttt{samehmetias@gmail.com} \\\AND
  Benjamin Roth \\
  Research Group Data Mining and Machine Learning \\
  University of Vienna \\
  Vienna, Austria \\
  \texttt{beroth@univie.ac.at} \\
}

\begin{document}
\maketitle


\begin{abstract}
We propose to use reinforcement learning to \emph{inform} transformer-based contextualized link prediction models by providing paths that are most useful for predicting the correct answer. 
This is in contrast to previous approaches, that either used reinforcement learning (RL) to directly search for the answer, or based their prediction on limited or randomly selected context.
Our experiments on WN18RR and FB15k-237 show that contextualized link prediction models consistently outperform RL-based answer search, and that additional improvements (of up to 13.5\% MRR) can be gained by combining RL with a link prediction model. 
The PyTorch implementation of the RL agent is available at \url{https://github.com/marina-sp/reinform}.
\end{abstract}

\vspace{1pt}
\section{Introduction}
\vspace{3pt}
In link prediction, also known as knowledge graph (KG) completion, the task is to find missing entries in a KG, based on other information already contained in the graph.
A common formulation of this problem is to present incomplete tuples of the form $(e_x, r_q, ?)$ for which a model is expected to find the missing entity $e_y$ that stands in relation $r_q$ to the entity $e_x$.
In this paper we tackle the task of \textit{contextualized} link prediction for KGs, where additional information about $e_x$ from its graph neighborhood, later referred to as \textit{context}, is engaged in the prediction process.

In contextualized link prediction, two strategies have been proposed: (1) \emph{Search-based}, were the answer entity is expected to be contained in an existing path, and relevant paths are searched for, based on the query tuple \cite{lao2011random,das-etal-2018-go}; (2) \emph{Prediction-based}, where the missing entity is predicted (out of all known entities) from a (contextualized) representation of the query tuple \cite{bordes2013translating,wang-etal-2019-coke}.

Minerva \cite{das-etal-2018-go} is a prominent neural \emph{search-based} approach for KG completion. 
In Minerva, a path search is performed using neural reinforcement learning (RL), and the entity at the endpoint of the returned path is taken as the answer to the query tuple.
CoKE \cite{wang-etal-2019-coke} is an highly effective neural \emph{prediction} architecture for contextualized link prediction in knowledge graphs.
CoKE takes chains of knowledge graph tuples and predicts missing entities using the Transformer architecture \cite{vaswani-2017-attention}.

In this work, we explore how to combine the advantages of both worlds, leveraging search to  provide the most useful information to a prediction model, which then has the freedom to predict any missing entity (even if not on a path returned by the search).
This is achieved through the interplay of two neural networks:
a prediction network, which bases its prediction on a path returned by the path search, and a path search that provides paths to the prediction model.
We compare two transformer architectures for the prediction model: \emph{Transform-CoKE}, that uses chains of relations only, as in CoKE, and \emph{Transform-InterEnt}, an extension of CoKE that includes intermediate entities in the paths. 
In addition, we integrate a RL architecture based on Minerva  as a path search model. 
However, in order to overcome the answer accessibility limitation of Minerva and tailor the search to best benefit the prediction model, the search model is trained with a modified loss function that takes the prediction model into account.

Experiments on FB15k237 \cite{toutanova-chen-2015-observed} and WN18RR \cite{dettmers-2017-conve} show that
RL-based path selection trained in combination with \emph{Transform-Coke} consistently yields better results than performing search only (Minerva) or than providing randomly sampled paths as context.
The \emph{Transform-InterEnt} extension to \emph{Transform-CoKE} 
performs better for FB15k237 but worse for WN18RR than \emph{Transform-CoKE}.

\section{Related Work}

Early approaches to representation-based link prediction were based on \emph{knowledge graph embeddings} \cite{bordes2013translating}. 
Predictions for $e_y$ are done solely based on learned vector representations of $r_q$ and $e_x$, which need to encode all relevant information.
(See the survey by \citet{rossi-etal-2021-knowledge} summarizing such context-free embedding approaches for link prediction.)
A more versatile approach is to employ neural models for \emph{contextualized} link prediction, which allows for utilizing and combining the information of a wider context around the query entity $e_x$.

DeepWalk \cite{perozzi-etal-2014-deepwalk} applied a language-modeling approach to paths in a graph, obtaining static node embeddings. Then, RNN-based models were used to incorporate context of entities from KGs and obtain self-sufficient deep contextualized embeddings \cite{das-etal-2017-chains, guo-etal-2019-learning}, as well as an additional component to context-free embeddings \cite{wang-etal-2018-dolores}. 

Ever since the introduction of the Transformer \cite{vaswani-2017-attention} and specifically BERT architecture \cite{devlin-etal-2019-bert}, a multitude of NLP tasks and other fields has been benefiting from these approaches. The power of contextualizing embeddings with the attention mechanism for KG was shown by \cite{wang-etal-2019-coke}, where the authors introduced the CoKE model.
To leverage non-linear context, graph convolution networks have also been applied to graph neighborhood of $e_x$ \cite{Shang_Tang_Huang_Bi_He_Zhou_2019, vashishth-2019-composition, bhowmik-2020-joint}.

Reinforcement learning has been exploited for the link prediction task, specifically for finding a path connecting a query entity with an answer \cite{xiong2017deeppath,das-etal-2018-go, godin-etal-2019-learning}. To the best of our knowledge, RL strategies have not yet been used to benefit a contextualized predictor.

\section{Overview}

\textbf{Minerva} \cite{das-etal-2018-go} is a RL-based approach to link prediction. It searches for paths in KG from a source entity to find and answer entity for a given incomplete query tuple.
The last entity of the most probable path is considered the prediction. 
For ranking-based evaluation, the top-N most probable paths according to the RL agent are generated for evaluation. Using this ranking, evaluation metrics such as Hits@k and MRR can be calculated in a usual fashion. An LSTM-encoded traversal history and node embeddings are used to learn a policy with a policy gradient method REINFORCE \cite{williams-2004-simpleSG}.

\textbf{CoKE} \cite{wang-etal-2019-coke} is a Transformer-based model for either embedding-based link prediction (trained on pure triples, not paths) or path query answering (PQA, considering longer paths). 
In PQA, the setting is that the model needs to recover the e4 from a path $e_1, r_1, r_2, r_3, e_4$, where $e_4$ is unacessible during prediction. 
PQA is essentially a multi-hop reasoning task, since the prediction is always made for the \textit{last element of the path} (an entity).
The problem specification in PQA relies on fixed prediction paths that cannot be changed. 
In contrast, our setting uses paths to enhance link prediction, i.e. fill in the \textit{missing position of the triple}, where the model has the freedom to find additional paths to support its decision (either through sampling on the fly, or through RL).


We use two datasets for our experiments: \textbf{FB15k-237} and \textbf{WN18RR}. The former is a subset of Freebase \cite{bollacker-2008-freebase}, a collection of facts about real-world entities (e.g. celebrities, locations, events), whereas the latter stems from WordNet \cite{miller-1992-wordnet} and contains semantic relations between lexical units of the English language.

\begin{center}
\begin{tabular}{lcc}
	\toprule 
	Dataset & \#entities & \#relations \\
	\midrule
	FB15k-237 & 14,541 & 237 \\
	WN18RR & 40,943 & 11 \\
	\bottomrule 
\end{tabular}
\end{center}
\begin{table*}
	\centering
	\begin{center}
		\begin{tabular}{ll c cc c cc}
			\toprule 
			\multicolumn{2}{l}{Model} & & \multicolumn{2}{c}{FB15k-237} & & \multicolumn{2}{c}{WN18RR} \\
			Name & N & & H@1 & MRR & & H@1 & MRR \\
			\midrule
			Minerva & 3 & & 0.1056 & 0.1516 & & 0.3618 & 0.3942 \\
			\midrule
			Transform-CoKE + sampling & 2 & & 0.1781 & 0.2427 & & 0.2208 & 0.2960 \\
			Transform-CoKE + Minerva  & 3 & & 0.1758 & 0.2344 & & \textbf{0.3816} & 0.4209 \\
			Transform-CoKE + RL & 2 & & \textit{0.2191} & \textit{0.2910} & & 0.3674 & \textit{\textbf{0.4312}} \\
			\midrule
			Transform-InterEnt + sampling & 3 & & 0.2238 & 0.3040 & & 0.2454 & 0.3065 \\			
			Transform-InterEnt + Minerva  & 3 & & \textbf{0.2242} & \textbf{0.3041} & & 0.2498 & 0.3095 \\
			Transform-InterEnt + RL & 3 & & 0.2241 & 0.3040 & &\textit{0.3036} & \textit{0.3552} \\ 
			\bottomrule 
		\end{tabular}
		\caption{Link prediction Hits@1 and MRR on \textbf{test} set for FB15k-237 and WN18RR. Bold denotes the best metric for a data set across all models, italic marks where our RL model yields best performance across different context generation strategies within a specific model variant for one data set.}
		\label{tab:test_scores}
	\end{center}
\end{table*}

\section{Experiments}
In our experiments, we vary how context paths are retrieved from the graph. 
These paths are then given to one of pretrained Transformer models for entity prediction:
\textit{Transform-CoKE} with middle entities omitted and \emph{Transform-InterEnt}, that uses the full path, including the middle entities.
For a triple $(e_x, r, e_y)$, a context path with $N$ relational steps $r_{xi}, e_{xi}$ starting from $e_y$ is produced according to the selected retrieval strategy. 
E.g., a unmasked input sequence for \emph{Transform-InterEnt} with context length $N=2$ then has the form $e_x, r_q, e_y, r_{x1}, e_{x1}, r_{x2}, e_{x2}$, with 2 steps $r_{xi}, r_{xi}$ taken from $e_y$.

We compare three context path generation strategies:
\textbf{sampling} generates a sampled context path (not informed by the query).
\textbf{Minerva} takes the most probable path that Minerva has taken for the query tuple.
\textbf{RL} obtains context returned by the RL agent trained in conjunction with the prediction model.
The following chapter formalizes our approach to RL-based context generation.







\subsection{Pretraining paths}
Let $D$ be a set of original triples of form $(e_x, r_q, e_y)$ where $e_x, e_y \in \mathcal{E}$ are entities and $r \in \mathcal{R}$ is a relation.
Similarly to Minerva, we introduce an inversed relation $r^{-1} \in \mathcal{R}^{-1}$ for every relation $r \in \mathcal{R}$ in order to provide the search models with access to all nodes during graph traversal.
A reversed triple is added for every triple during both training and evaluation, resulting in an extended set $D' = D \cup \{(e_y, r_q^{-1}, e_x) \vert (e_x, r_q, e_y) \in D \}$.
This way, the first position is always the masked one (for head or tail prediction), while context generation starts from the last position for any triple from $D'$. Pretraining paths are sampled randomly from a graph constructed from the train triples. For a detailed description of sampling process see Appendix A.

As a result, a set of yet unmasked chains is obtained, with the following format: $c = (e_x, r_q, e_y, r_{x1}, e_{x1}, \dots, r_{xN}, e_{xN}) $ that are used directly as input for \textit{Transform-InterEnt}. For \textit{ Transform-CoKE}, middle entities are omitted to fit the expected input structure of a CoKE model $c = (e_x, r_q, r_{x1}, \dots, r_{xN}, e_{xN})$. 

\subsection{Pretraining of contextualized predictors}
In its essence, the task of link prediction is equivalent to that of masked language modeling: the model learns to recover a masked element in a sequence of items stemming from a limited vocabulary, specifically the first entity is masked to then be predicted in a sampled chain $c$ from KG.
Despite that only entities appear in the masked positions and should be predicted, the
predictor's vocabulary comprises\\
 $V = \{\mathcal{E} \cup \mathcal{R} \cup \mathcal{R}^{-1} \}$ both entities and 
relations\footnote{The vocabulary further includes BERT-specific tokens (MASK, CLS, UNK and SEP) that are omitted here.} as they are treated as equal elements of a sequence (same as nouns and verbs are not separated in BERT). 
Same vocabulary is used in the RL-search component to allow for a direct use of its output paths as input to the predictor.\footnote{To account for compatibility of Minerva-generated paths with the transformer predictor, we use the UNK token as a NO\_OP (no operation, stay in the current graph node) token.}

In the pretraining phase, the scorer is optimized to correctly recover a masked entity for a sampled path. Let $\mathbf{h}^j_k(c) \in \mathbb{R}^{d}$ be the hidden Transformer representation of the \textit{k-th} position in the \textit{j-th} layer obtained for an input chain $c$.
Then, the pretraining objective can be written as \\
\vspace{2pt} \hspace{5pt} $L(c) = CrossEntropy(PredHead(\mathbf{h}^L_0(c)), c_0)$,\\
where $L$ is the last encoding layer, $c$ is an unmasked input path (chain) with the expected entity $c_0$ in the first position, and \textit{PredHead} is a one of the predictor-specific final decision layers \textit{PredHead}: $\mathbb{R}^{d} \mapsto \mathbb{R}^{|V|}$, following the source implementations.\footnote{For \textit{Transform-InterEnt}, the HuggingFace \cite{wolf-etal-2020-transformers} implementation for BERT is used, \textit{PredHead} corresponds to an \texttt{BertOnlyMLMHead}; for \textit{Transform-CoKE}, it is the last \textit{FF} layer in the original terminology.}
Hyper-parameter choices, such as the number of layers and the number of epochs, are modeled after the PQA setup in the CoKE paper.\footnote{Adjusting the hyperparameters to follow the setup of link prediction model, i.e. increasing the number of Tranformer layers and epochs, did not yield better performance.} 

\begin{figure}[h!]
	\includegraphics[width=0.95\linewidth]{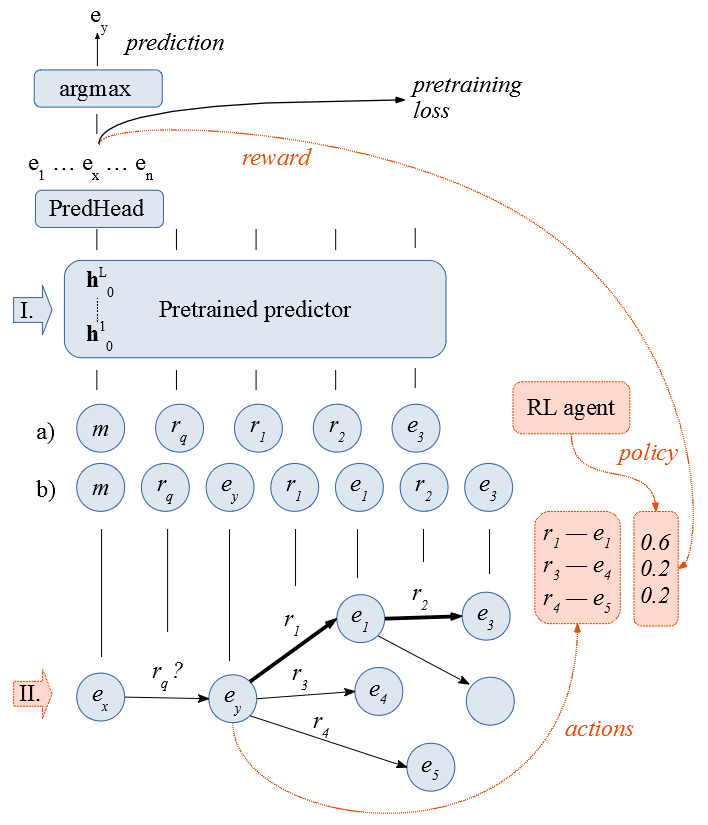}
	\caption{Workflow of the contextualized scorer component (I.) and the RL context-search (II.). Bold arrows represent the edges selected by the RL agent. $r_q$ marks the query triple, $m$ stands for the masked token. a) and b) stand for two path representations used by \textit{Transform-CoKE} and \textit{Transform-InterEnt} respectively.}
	\label{fig:schema_joint_architecture}
\end{figure}

\subsection{Training of the RL context selection}
On a high level, an RL agent selects the most probable \textit{action} according to a learned \textit{policy}. A positive feedback (a \textit{reward}) reinforces a beneficiary choice of the agent by making all actions that led to a rewarded state more probable in the future.
In our case, the RL agent chooses which step in a KG to take next based on the current position, previous steps taken and the query.
It has the same structure and hyper-parameters as the one in Minerva (based on our reimplementation)\footnote{
	See Section 2 of \cite{das-etal-2018-go} for a formal description of the agent. We were able to reproduce the results reported in \cite{das-etal-2018-go} with our reimplementation. 
	In the original Minerva paper, however, only $(e_x,r_q,?)$ queries are used for evaluation, i.e., it is only attempted to predict the object, see Section 3.1 "KB query answering" of \cite{das-etal-2018-go}.
	In our experiments we evaluate on subject and object positions, and Minerva results drop significantly in this more challenging setting.},
except for the reward function that now aims for a good contextualized prediction, rather than  finding the correct answer. 

The RL agent is trained separately, and uses the already pretrained transformer during training.\footnote{We also experimented with jointly optimizing the prediction and ML module, but ran into stability issues, yielding worse results.} 
The modified reward of the RL agent can be characterized using following terminology:
$S_t=(e_t, r_q, e_y)$ is the current state of the agent, where $e_t$ is the entity at time step $t$, $r_q$ is the relation and $e_y$ is the tail entity of the query; $c_t$ is the chain of KG-steps traversed by the agent.
The reward at final time step $N$ then equals\\
\vspace{2pt} \hspace{10pt} $R(S_N) = softmax(PredHead(\mathbf{h}^L_0(c_N)))_i$,\\ 
where \textit{PredHead} returns the logits over all possible entities of a deep scorer and $i$ is the vocabulary index of the correct answer entity $e_x$. The summary of the joint architecture is illustrated in Figure \ref{fig:schema_joint_architecture}. Our PyTorch implementation of the RL agent is available at \url{https://github.com/marina-sp/reinform}.

\subsection{Results and Analysis}
We evaluate the models with mean reciprocal rank (MRR) and Hits@1.
Table \ref{tab:test_scores} shows that the prediction-based approach generally outperforms the search-based approach of Minerva for FB15k-237.
Which path generation mechanism is used makes as marked difference for \textit{Transform-CoKE}, and training the RL model specifically for this setting performs best by a large margin.
Including intermediate entities in the path processed by the transformer (\textit{Transform-InterEnt}) again increases the performance for all path search strategies (but the differences between them disappear).

Using learned context paths over random sampling is generally beneficial in case of WN18RR. With a path for a test query \texttt{(tog VB1, derivationally related form$^{-1}$, dresser NN3)} extracted by the RL agent \texttt{(dresser NN3, derivationally related form$^{-1}$, get dressed VB1, derivationally related form, dresser NN3)} the correct prediction was scored the highest by the \textit{Transform-CoKE}, whereas with a randomly sampled one \texttt{(dresser NN3, hypernym, supporter NN3, hypernym$^{-1}$, hatchet man NN2)}, the expected entity \texttt{tog VB1} was ranked 2267.
The RL-path also exemplifies the ability of the model to generalize beyond the entities contained in a path, which Minerva can not per definition.

\section{Conclusion}

We have shown how to combine path selection by reinforcement learning with transformer-based prediction models for contextualized link prediction in knowledge graphs.
This approach achieves strong perfomance gains over a recent previous RL model that directly searches for an answer in the graph.
Analysis indicates that this performance gain is presumably due to the fact that answer entities often do not lie on paths found by RL, and need instead be predicted from the entire pool of entities (not constrained to entities reached on a path).
Our method also shows gains over using the transformer-based prediction models on paths randomly selected from around the query entity.
This shows the potential of reinforcement learning to benefit prediction models that rely on path selection.

\section*{Acknowledgements}
This work was funded by the Deutsche Forschungsgemeinschaft (DFG, German Research
Foundation) - RO 5127/2-1.

\bibliography{anthology,custom}
\bibliographystyle{acl_natbib}

%
%
%

\section*{Appendix}


\subsection*{A. Pretraining path sampling}
\label{app:pretraining}
For \textit{Transform-InterEnt}, a pretraining path, or \textit{chain} $c$, with a fixed amount of steps $K$ is sampled by randomly travesing the graph starting from $e_y$. A step from $e_y$ is a single outgoing edge described by the labels of the respective edge and target node $(r_{xi}, e_{xi})$. The same $N = K$ is used when retrieving the context with an RL agent.
The underlying graph consists of triples from the training set alone. 
The query triple $(e_x, r, e_y)$ itself as well as the backward connection $(e_y,r^{-1},e_x)$ are excluded from the sampling process to resemble the evaluation process, where the query triple is not available during graph traversal. 
The described process is equvalent to the \textit{sampling} strategy.

For \textit{Transform-CoKE}, the set of pretraining paths has mixed lengths $1 <= K <= 5$ following the original implementation. 
The length of context $N$ during evaluation, i.e. the number of steps taken by the RL agent, is however constant. 
For both \textit{Transform-InterEnt} and \textit{Transform-CoKE}, $N$ is treated as a hyper-parameter that is determined based on the development data.
\end{document}